\documentclass{article}

\usepackage{microtype}
\usepackage{graphicx}
\usepackage{subcaption}
\usepackage{booktabs} 

\usepackage{hyperref}

\usepackage[preprint]{icml2026}
\hypersetup{pdfsubject={Preprint}}

\usepackage{amsmath}
\usepackage{amssymb}
\usepackage{mathtools}
\usepackage{amsthm}

\usepackage[capitalize,noabbrev]{cleveref}

\theoremstyle{plain}

\theoremstyle{definition}

\theoremstyle{remark}

\usepackage[textsize=tiny]{todonotes}

\begin{document}

\twocolumn[
  \icmltitle{Asymmetry Tunneling Beyond Adaptive Equivariance Relaxation}



  \icmlsetsymbol{equal}{*}

  \begin{icmlauthorlist}
    \icmlauthor{Eunho Jeong}{snu}
    \icmlauthor{Joon Jang}{snu}
    \icmlauthor{Hyeonjin Kim}{snu}
  \end{icmlauthorlist}

  \icmlaffiliation{snu}{Seoul National University}

  \icmlcorrespondingauthor{Eunho Jeong}{ckddm324@gmail.com}
  \icmlcorrespondingauthor{Hyeonjin Kim}{rachmaninov4ever@gmail.com}

  \vskip 0.3in
]


\printAffiliationsAndNotice{}  

\begin{abstract}
Equivariant neural networks encode prescribed symmetries as powerful inductive biases, yet real-world data may exhibit symmetry-breaking structure that these priors cannot fully capture.
Prior work has largely addressed this challenge through model-side equivariance relaxation, seeking an appropriate level of equivariance for the data.
This raises a complementary question: must equivariance relaxation be adaptively controlled?
To investigate this question, we introduce a residual path alongside a strictly equivariant model and deliberately restrict their interface to examine the transmission of task-relevant structure.
Across five benchmarks, we observe both residual collapse and survival despite these restrictions; we call the latter phenomenon asymmetry tunneling.
We further characterize this phenomenon through four complementary diagnostics. Our results suggest that adaptive control of equivariance relaxation is not a prerequisite for capturing potential symmetry-breaking structure.
\end{abstract}

\section{Introduction}
Equivariant neural networks \cite{introexam_cohen_GECN,introexam_Ravanbakhsh_Equi,introexam_kondor_2018_generalization}, which encode prescribed symmetries to improve sample efficiency and generalization \cite{relatedworkchen2020groupthro,relatedworkbietti2019groupinv}, have been widely applied in computer vision \cite{introexam_cohen_2017_steerable}, graph learning \cite{introexamgilmer2017neural}, and 3D modeling \cite{introexamschutt2017schnet}.
However, real-world data can contain structure that is not fully captured by a strictly equivariant prior \cite{main_disco_smidt2021finding,main_imperfect_wang2022approximately,introexamwang2023discovering}, due to noise, measurement bias, chirality, local interactions, or other symmetry-breaking effects.
To address imperfect symmetries in real-world data, prior work has explored soft equivariance priors, relaxed equivariant architectures, theoretical analyses of approximation--generalization trade-offs, and data-driven adaptation \cite{main_RPP_finzi2021residualpath,main_imperfect_wang2022approximately,main_tradeoff_petrache2023approximation,main_ACE_manolache2025learning}.
These works examine model-side equivariance relaxation, reflecting the view that model equivariance should be adapted to an appropriate level for the data.

This pursuit of an appropriate level of equivariance raises a complementary diagnostic question: \emph{must equivariance relaxation be adaptively controlled to capture symmetry-breaking structure?}
To investigate this question, we use a strictly equivariant model as a reference anchor and introduce an expressive residual path through a deliberately restricted interface.
We then test whether this restricted residual path can transmit task-relevant contributions.
The residual path is designed with ample expressive capacity to represent potential symmetry-breaking structure, while the restricted interface uses stop-gradient, scalar gating, and a single additive fusion to limit residual--anchor interaction and transmission.
Empirically, the restricted residual exhibits two regimes: collapse and survival.
We call residual survival despite these restrictions \emph{asymmetry tunneling}.
To characterize this phenomenon, we employ four complementary empirical diagnostics: \emph{residual survival}, measured relative to the anchor; \emph{anchor-relative geometry}, by comparing the residual with local variation in anchor representations; \emph{physical localization}, by tracing its concentration in physical structure; and \emph{task relevance}, through held-out performance.
Our contributions are as follows:
\begin{enumerate}
    \item We identify \emph{asymmetry tunneling} as residual survival despite a restricted interface with a strictly equivariant anchor, implemented through stop-gradient, scalar gating, and a single additive fusion. Across tasks, we observe both residual survival and collapse.
    \item We characterize these patterns through complementary empirical diagnostics of residual survival, anchor-relative geometry, physical localization, and task relevance to probe potential symmetry-breaking structure.
\end{enumerate}

\section{Related Work}
Equivariant neural networks encode prescribed group symmetries directly into model architectures
\cite{introexam_cohen_GECN,introexam_Ravanbakhsh_Equi,introexam_cohen_2017_steerable,introexam_kondor_2018_generalization}
and can improve generalization and sample efficiency when the imposed symmetry matches the task
\cite{relatedworkchen2020groupthro,relatedworkbietti2019groupinv,relatedworksannai2021improvedgener}.
Representative geometric architectures include EGNN
\cite{relatedworkEGNNsatorras2021n},
SEGNN
\cite{main_SEGNN_brandstetter2021geometric},
and Vector Neurons
\cite{main_VN_deng2021vector}.
Applications span 3D point-cloud models such as VN-DGCNN and TetraSphere
\cite{relatedworkDGCNNwang2019dynamic,main_VN_deng2021vector,relatedworkmelnyk2024tetrasphere},
equivariant trajectory models such as EGNO and EqMotion
\cite{main_EGNO_xu2024equivariant,main_EqMotion_xu2023eqmotion},
and GVP-based models for molecular structure
\cite{main_gvp_jing2020learning,main_gvp_atom3d_jing2021equivariant}.
Real-world data, however, can contain structure that is not fully captured by a prescribed equivariant prior, including contact-dependent motion, cluttered or partially occluded point clouds, and molecular chirality
\cite{main_CMU_mao2022contact,main_scanobjectNN_uy2019revisiting,main_cannotlearnchirality_dumitrescu2024}.

Prior work has examined how model-side equivariance can be relaxed to better reflect the symmetry structure of the data, alongside theoretical analyses of model--data symmetry alignment.
Residual Pathway Priors soften hard equivariance constraints into priors biased toward equivariant solutions
\cite{main_RPP_finzi2021residualpath};
approximately equivariant networks relax weight sharing or tying
\cite{main_imperfect_wang2022approximately};
theoretical work studies the relation between model and data equivariance error
\cite{main_tradeoff_petrache2023approximation};
and ACE controls deviations from equivariance through data-driven constrained training
\cite{main_ACE_manolache2025learning}.
A related line of work seeks to discover symmetry structure from data and incorporate it into the model through latent groups, equivariance-inducing parameterizations, symmetry-breaking inputs, Lie generators, generative transformations, or relaxed weights
\cite{main_disco_zhou2020meta,main_disco_smidt2021finding,main_disco_dehmamy2021automatic,pl_desai2022symmetry,main_disco_yang2023generative,introexamwang2023discovering}.

Our work takes a complementary diagnostic perspective.
Rather than adaptively controlling equivariance relaxation, we use a deliberately restricted interface with a strictly equivariant anchor to examine residual survival and collapse.
We further examine surviving residuals as diagnostic signals of potential symmetry-breaking structure.

\begin{figure}[htbp]
  \centering
  \includegraphics[width=1\linewidth]{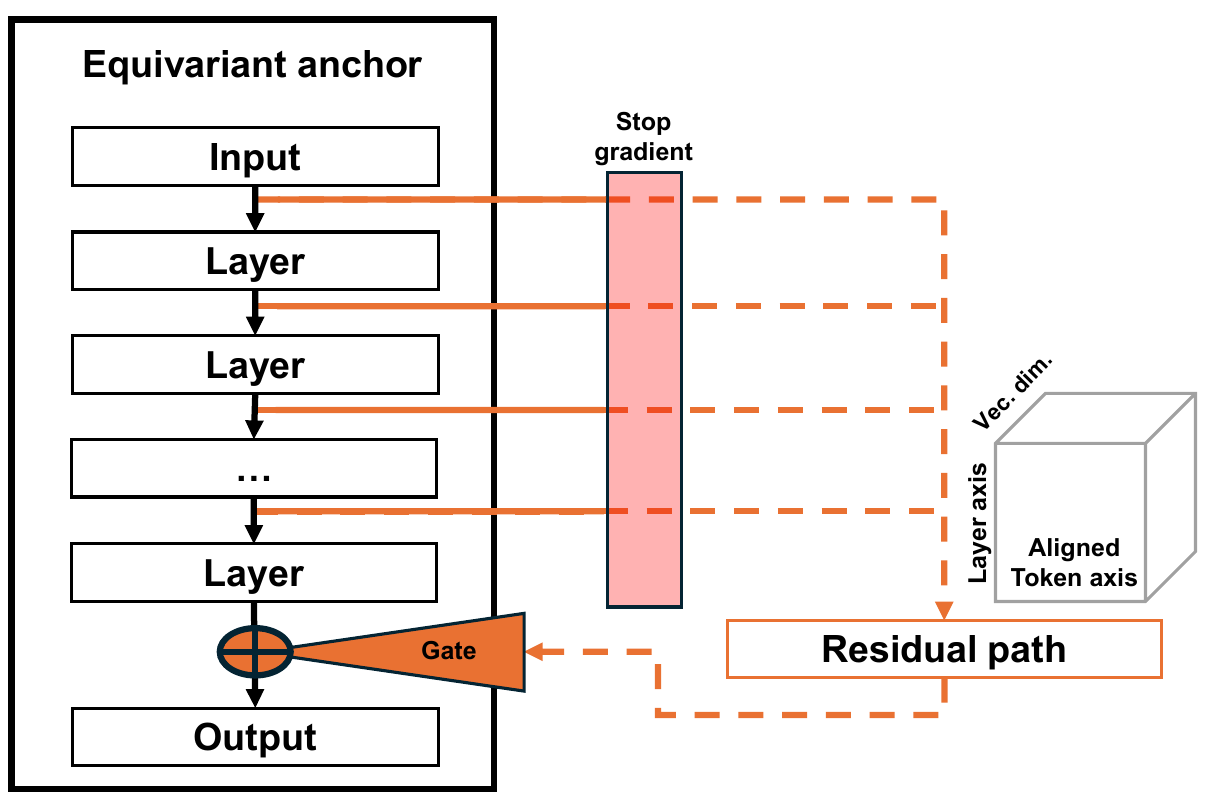}
\caption{Overview of the restricted residual architecture. The residual path reads detached anchor-layer representations, and its gated contribution is fused once after the final anchor layer.}
  \label{fig:main_overview}
\end{figure}

\section{Residual Architecture}
\label{sec:architecture}
To diagnose residual transmission beyond a strictly equivariant anchor, we restrict residual–anchor interaction while keeping the transmitted contribution measurable and physically traceable. Figure~\ref{fig:main_overview} illustrates this restricted residual interface.

\paragraph{Notation and architecture.}
Let $x\in\mathcal{X}$, let $G$ denote the symmetry group acting on $\mathcal{X}$, and let $\theta$ denote the anchor parameters.
We denote the anchor representation at layer $\ell$ by
$h_\ell(x)=(h_{\ell,1}(x),\ldots,h_{\ell,P}(x))\in\mathcal{Z}$,
where $\mathcal{Z}:=\mathbb{R}^{PC}$ is the vectorized representation space.
Here $P$ is the number of latent elements, referred to as tokens,
$C$ is the feature dimension of each token, and
$\ell=0,\ldots,L$ indexes anchor layers.
Each $h_{\ell,p}(x)\in\mathbb{R}^C$, where $p=1,\ldots,P$.
We denote the full latent history by
$H_\theta(x)=\{h_{\ell,p}(x)\}_{\ell,p}$
and define the final anchor representation as
$m_\theta(x):=h_L(x)$, which satisfies
\begin{equation}
    m_\theta(g\!\cdot\!x)=\rho(g)m_\theta(x),
    \qquad g\in G,
    \label{eq:strict_anchor}
\end{equation}
where $\rho(g)$ denotes the corresponding action on $\mathcal{Z}$.
The residual path $\mathcal{R}_\psi$, parameterized by $\psi$,
reads the anchor history through stop-gradient
$\operatorname{sg}(\cdot)$ and produces a token-aligned residual
proposal $s_\psi(x):=\mathcal{R}_\psi(\operatorname{sg}(H_\theta(x)))\in\mathcal{Z}$.

A non-learnable scalar gate $\lambda_t>0$, initialized with $\lambda_0>0$, is updated from the loss trend and held constant within each optimization step.
Given the training loss $L_t$, we track its exponential moving average $\bar L_t=\beta\bar L_{t-1}+(1-\beta)L_t$. The loss-trend-based gate update can be summarized as
\begin{equation}
\lambda_t=\lambda_{t-1}
\left(
1-\eta\frac{\bar L_{t-1}-\bar L_t}{\bar L_{t-1}}
\right),
\label{eq:gate_update}
\end{equation}
where \(0\leq\beta<1\) is the EMA decay and \(0<\eta<1\) controls the gate response.
Thus, a decreasing loss trend progressively suppresses residual transmission.
Accordingly, the residual transmitted to the fusion point is the gated proposal, and fusion occurs exactly once:
\begin{equation}
r(x):=\lambda_t s_\psi(x)\in\mathcal{Z},
\quad
z(x)=m_\theta(x)+r(x)\in\mathcal{Z}.
\label{eq:residual_fusion}
\end{equation}
Here $s_\psi$ denotes the proposed residual computed from the detached anchor history, while $r$ denotes the transmitted residual; both are represented in the same latent space $\mathcal{Z}$ as the anchor $m_\theta$ and fused state $z$.

\subsection{Restricted Residual Interface}
\label{sec:restricted_interface}
The interface restricts residual--anchor interaction and residual transmission along three distinct axes: backward gradient flow, forward interaction, and transmitted residual magnitude.
\paragraph{Backward restriction: stop-gradient under additive fusion.}
Stop-gradient satisfies
\begin{equation}
    \operatorname{sg}(u)=u,
    \qquad
    D_u\operatorname{sg}(u)=0.
    \label{eq:sg_definition}
\end{equation}
For a general fusion $z(x)=F(m_\theta(x),r(x))$, stop-gradient gives
$D_\theta\operatorname{sg}(H_\theta(x))=0$, and hence $D_\theta r(x)=0$. The anchor-side derivative therefore reduces to
\begin{equation}
    D_\theta z(x)
    =
    D_mF\,D_\theta m_\theta(x).
    \label{eq:sg_general_fusion}
\end{equation}
For the additive fusion in \cref{eq:residual_fusion}, $D_mF=I$, yielding
\begin{equation}
    D_\theta z(x)=D_\theta m_\theta(x).
    \label{eq:sg_additive_fusion}
\end{equation}
Thus, stop-gradient and additive fusion jointly restrict residual-mediated
gradient flow to the anchor while keeping the anchor-side Jacobian equal to
$D_\theta m_\theta(x)$.

\paragraph{Forward restrictions: single fusion and scalar transmission.}
This backward restriction does not constrain forward interaction.
To restrict forward interaction, we fuse the residual with the anchor only once, after the final anchor layer, preventing repeated interaction across anchor layers.
As the smoothed loss decreases, the scalar gate $\lambda_t$ in Equation~\eqref{eq:gate_update} progressively suppresses residual transmission.
Together, these mechanisms impose forward restrictions: single fusion prevents repeated interaction across anchor layers, while scalar gating limits residual transmission magnitude.

\subsection{Measurable and Traceable Residual Representation}
\label{sec:measurable_residual}
While these restrictions limit residual transmission, we also design the residual representation to make its contribution measurable and physically traceable.

\paragraph{Shared latent space for anchor-relative geometry.}
The anchor, transmitted residual, and fused representation lie in the same
latent space $\mathcal{Z}$. We denote the set of representations produced by
the anchor as
\begin{equation}
    \mathcal{A}_\theta
    :=
    \{m_\theta(x):x\in\mathcal{X}\}
    \subset \mathcal{Z}.
    \label{eq:anchor_representation_set}
\end{equation}
Since
\(
r(x)=z(x)-m_\theta(x),
\)
the transmitted residual defines a displacement from the anchor state
$m_\theta(x)$ within $\mathcal{Z}$.
Crucially, since $r(x)$ need not be a displacement between
states in $\mathcal A_\theta$, the fused state
$z(x)$ may lie outside $\mathcal A_\theta$.
To characterize this geometric distinction, let
$\widehat T_{m_\theta(x)}(\mathcal A_\theta)$ denote a local
linear approximation to the anchor variation around
$m_\theta(x)$, and define the approximate normal space as
$\widehat N_{m_\theta(x)}(\mathcal A_\theta)
:=(\widehat T_{m_\theta(x)}(\mathcal A_\theta))^\perp$.
The transmitted residual admits the orthogonal decomposition
\begin{equation}
\begin{gathered}
r(x)=r^\parallel(x)+r^\perp(x),\\
r^\parallel(x)\in\widehat T_{m_\theta(x)}(\mathcal A_\theta),\quad
r^\perp(x)\in\widehat N_{m_\theta(x)}(\mathcal A_\theta).
\end{gathered}
\label{eq:residual_orthogonal_decomposition}
\end{equation}
The transverse component $r^\perp(x)$ captures residual variation orthogonal to the approximated local anchor tangent space. 
This allows us to assess whether residual survival is accompanied by geometrically distinct contributions in the latent representation space, beyond local anchor variation.

\paragraph{Layer-history aggregation with token-aligned assignment.}
Single fusion leaves two requirements for the residual.
It should access information from intermediate anchor layers rather than relying only on $h_L$, while preserving token-aligned residual assignments for physical tracing despite aggregating information across layers and tokens.
To satisfy these requirements, we use Transformer-style cross-attention in which the detached final anchor states $h_{L,p}$ serve as queries and the memory is constructed from the detached layer--token history.
Each final-layer query $h_{L,p}$ aggregates information across the full memory into a single residual contribution $s_{\psi,p}$ while retaining token index $p$:
\[
    s_{\psi,p}(x)
    =
    R_{\psi,p}\!\left(
        \operatorname{sg}\!\left(H_\theta(x)\right)
    \right),
    \qquad p=1,\ldots,P.
\]
Thus, the memory may integrate information across layers and tokens, while $s_{\psi,p}$ remains assigned to query token $p$.
When token index $p$ corresponds to a physical element, the token-aligned proposal $s_{\psi,p}$ preserves this correspondence in the transmitted residual $r_p$.
This allows us to assess whether residual survival is accompanied by localization to specific physical elements.

\section{Empirical Residual Diagnostics}
\label{sec:diagnostics}
Building on the measurable and physically traceable residual representation in Section~\ref{sec:measurable_residual}, for each evaluation sample $x_i\in\mathcal D$, we define $m_i:=m_\theta(x_i)$ and $r_i:=r(x_i)$ as the anchor and transmitted residual representations.
We evaluate four empirical diagnostics: \emph{survival}, \emph{anchor-relative geometry}, \emph{physical localization}, and \emph{task relevance}.

\paragraph{Residual survival.}
We measure whether the residual entering fusion remains non-negligible relative to the anchor:
\begin{equation}
    \kappa_r
    =
    \frac{
        \operatorname{RMS}_{i\in\mathcal D}\!\left(r_i\right)
    }{
        \operatorname{RMS}_{i\in\mathcal D}\!\left(m_i\right)
    }.
    \label{eq:residual_survival}
\end{equation}
Here RMS denotes the root-mean-square over the evaluation set.
We use empirical thresholds of $\kappa_r < 0.05$ for collapse, $0.05 \leq \kappa_r < 0.1$ for weak survival, and $\kappa_r \geq 0.1$ for survival.
In surviving cases, the residual proposal $s_\psi$ increases in magnitude as the scalar gate $\lambda_t$ decreases, compensating for the attenuation and sustaining residual transmission.

\subsection{Anchor-Relative Local Geometry}
\label{sec:diagnostic_geometry}
We measure whether surviving residuals contribute to the fused representation in geometric directions beyond the anchor's local variation, building on the approximate tangent--normal decomposition in Section~\ref{sec:measurable_residual}.
Using an independent anchor reference bank, let $\{m_{j_1},\ldots,m_{j_K}\}$ be the Euclidean $K$-nearest neighbors of $m_i$.
We apply singular value decomposition (SVD) to their local differences:
\begin{equation}
    \begin{bmatrix}
        (m_{j_1}-m_i)^\top\\
        \vdots\\
        (m_{j_K}-m_i)^\top
    \end{bmatrix}
    =
    U_i\Sigma_iV_i^\top,
    \,\,\,
    P_i
    =
    V_{i,1:d_i}V_{i,1:d_i}^{\top},
    \label{eq:diagnostic_local_svd}
\end{equation}
where \(d_i\) is selected using a prespecified singular-energy threshold \(\delta\).
The projector $P_i$ provides a Local-SVD approximation to the tangent--normal decomposition in Equation~\eqref{eq:residual_orthogonal_decomposition}:
\begin{equation}
\operatorname{Im}(P_i)\approx\widehat T_{m_i}(\mathcal A_\theta),\quad
\operatorname{Im}(I-P_i)\approx\widehat N_{m_i}(\mathcal A_\theta).
\end{equation}
Accordingly, $(I-P_i)r_i$ estimates the transverse component of the transmitted residual in this space.
We measure its energy fraction by
\begin{equation}
    \rho_{\mathrm{res},i}^{\perp}
    =
    \frac{
        \|(I-P_i)r_i\|_2^2
    }{
        \|r_i\|_2^2
    }.
    \label{eq:residual_transverse_energy}
\end{equation}
To calibrate this transverse energy fraction against local anchor variation, we reserve the next $H$ nearest anchor representations as held-out controls.
For the next \(H\) nearest anchor representations, define \(a_{ih}:=m_{j_{K+h}}-m_i\), \(h=1,\ldots,H\). The excess transverse energy is then
\begin{equation}
    \Delta\rho_i^\perp
    =
    \rho_{\mathrm{res},i}^{\perp}
    -
    \frac{1}{H}
    \sum_{h=1}^{H}
        \frac{
            \|(I-P_i)a_{ih}\|_2^2
        }{
            \|a_{ih}\|_2^2
        }.
    \label{eq:excess_transverse_energy}
\end{equation}
A larger $\Delta\rho_i^\perp$ indicates greater geometric transversality of the transmitted residual relative to held-out anchor variation in the latent representation space.
For readability, reported $\Delta\rho^\perp$ values are multiplied by $100$ and shown as percentage-point differences.

\subsection{Physical Localization and Task Relevance}
\label{sec:diagnostic_physical_task}
We evaluate whether surviving residuals localize to task-relevant physical structure, building on the token-aligned residual representation in Section~\ref{sec:measurable_residual}.
We distinguish two tracing approaches: direct element-wise tracing with preserved token correspondence and attribution-based tracing after global aggregation.

\paragraph{Attribution-based tracing.}
When global aggregation in the task-specific readout, such as pooling, removes explicit token--physical-element correspondence, residual influence cannot be localized directly to physical elements.
We therefore localize residual influence to physical elements using Grad-CAM \cite{expselvaraju2017grad}.
For each evaluation sample $x_i\in\mathcal D$, let $\mathcal{T}_i$ denote the selected high-attribution elements, $\mathcal{B}_i$ a predefined subset of input physical elements, and $N_i$ the total number of physical elements.
We quantify localization by the lift
\begin{equation}
    \operatorname{Lift}_i(\mathcal{B}_i)
    =
    \frac{
        |\mathcal{T}_i\cap\mathcal{B}_i|/|\mathcal{T}_i|
    }{
        |\mathcal{B}_i|/N_i
    }.
    \label{eq:diagnostic_lift}
\end{equation}
The lift is the ratio of the observed overlap between Grad-CAM high-attribution elements and the predefined physical structure to the overlap expected under uniform selection.
A larger $\operatorname{Lift}_i(\mathcal{B}_i)$ indicates stronger localization to the predefined physical structure, with $\operatorname{Lift}=1$ corresponding to chance-level localization.

\paragraph{Direct element-wise tracing.}
When token--physical-element correspondence is preserved, the transmitted residual can be localized directly to physical elements by element-wise tracing.
Let $r_{i,p}$ denote the slice of $r_i$ corresponding to physical element $p$.
To isolate variation across physical elements, we center each slice by the sample-wise mean $\bar r_i:=P^{-1}\sum_{p=1}^{P} r_{i,p}$.
We then measure direct element-wise localization by
\begin{equation}
    L_p
    =
    \frac{1}{|\mathcal D|}
    \sum_{i\in\mathcal D}
    \frac{
        \|r_{i,p}-\bar r_i\|_2^2
    }{
        \|r_{i,p}\|_2^2
    }.
    \label{eq:diagnostic_direct_localization}
\end{equation}
Larger $L_p$ indicates stronger localization across physical elements.

\paragraph{Task relevance.}
We assess task relevance by comparing held-out performance between the restricted-residual model and an independently trained strictly equivariant baseline without the residual path.
Improved held-out performance supports the residual's task relevance.

\subsection{Diagnostic Summary}
\label{sec:diagnostic_summary}
The four empirical axes introduced above are measured by complementary observables.
Residual survival is measured by $\kappa_r$, anchor-relative geometry by the excess transverse energy $\Delta\rho^\perp$, physical localization by $\operatorname{Lift}$ when token--physical-element correspondence is removed or $L_p$ when it is preserved, and task relevance by held-out performance improvement.
Together, these diagnostics distinguish
whether the transmitted residual survives, differs geometrically from local anchor variation, localizes to physical structure, and contributes to held-out prediction.

\section{Experiments}
\label{sec:experiments}
We focus on real-world datasets that may naturally exhibit symmetry-breaking structure.
In contrast, we include Charged N-body as a synthetic control where the prescribed symmetry is expected to hold.
Our real-world experiments cover ETH/UCY pedestrian forecasting, CMU human motion forecasting, ScanObjectNN point-cloud classification, and PDBbind molecular affinity prediction.
Task-specific residual token correspondence for these benchmarks is detailed in Appendix~\ref{app:token_structure}.
We analyze residual behavior using the diagnostics in Section~\ref{sec:diagnostics}, following standard dataset splits and baseline task protocols.

\begin{table}[t]
\centering
\caption{
Residual survival and task performance on Charged N-body and ETH/UCY.
Charged N-body uses SEGNN~\cite{main_SEGNN_brandstetter2021geometric}
as the equivariant anchor and reports mean squared error (MSE).
ETH/UCY uses EqMotion~\cite{main_EqMotion_xu2023eqmotion}
and reports average/final displacement error (ADE/FDE; lower is better).
$\kappa_r$ measures residual survival as defined in
Equation~\eqref{eq:residual_survival}.
}
\label{tab:collapse}
\small
\begin{tabular}{lccc}
\toprule
Setting & Anchor & + ours & $\kappa_r$ \\
\midrule
N-body
    & 0.00625 & 0.00626 & 0.0080 \\
\midrule
ETH
    & 0.40/0.61 & 0.40/0.58 & 0.0970 \\
Hotel
    & 0.12/0.18 & 0.16/0.23 & 0.0697 \\
Univ
    & 0.23/0.43 & 0.24/0.43 & 0.0644 \\
Zara1
    & 0.18/0.32 & 0.20/0.34 & $7.9{\times}10^{-5}$ \\
Zara2
    & 0.13/0.23 & 0.12/0.21 & 0.0966 \\
\bottomrule
\end{tabular}
\end{table}

\subsection{Charged N-body}
\label{sec:nbody}
We evaluate the Charged N-body system
\cite{expnbodykipf2018neural} with SEGNN
\cite{main_SEGNN_brandstetter2021geometric} as a synthetic control.
Because the prescribed symmetry is expected to hold in the synthetic N-body setting \cite{expnbodypertigkiozoglou2024improving,main_ACE_manolache2025learning}, the restricted residual is expected to collapse.
In Table~\ref{tab:collapse}, we observe the expected residual collapse ($\kappa_r=0.0080$) with no improvement in predictive performance.
This residual collapse is consistent with the symmetry-matched nature of N-body.

\subsection{ETH/UCY}
\label{sec:ETH}
We evaluate ETH/UCY pedestrian trajectory prediction \cite{expeth34lerner2007crowds,expeth54pellegrini2009you} using EqMotion \cite{main_EqMotion_xu2023eqmotion} as the anchor under the standard setting \cite{expeth1alahi2016social,expeth20gupta2018social,expeth76yuan2021agentformer}.
As shown in Table~\ref{tab:collapse}, residual behavior varies across ETH/UCY subsets under the same EqMotion-based architecture and restricted residual interface.
ETH, Hotel, Univ, and Zara2 exhibit weak survival ($\kappa_r=0.064$--$0.097$), whereas Zara1 exhibits collapse ($\kappa_r=7.9\times10^{-5}$), with both improvements and degradations in average/final displacement error (ADE/FDE) across subsets. 
Overall, ETH/UCY exhibits a weak-survival regime with subset-dependent collapse and inconsistent changes in predictive performance.

\subsection{CMU Motion}
\label{sec:CMU}
We evaluate the CMU motion ``Run'' and ``Walk'' datasets \cite{expcmu2003mocap}, using EGNO with the same settings \cite{main_EGNO_xu2024equivariant} as the anchor.
CMU exhibits asymmetry tunneling, which we characterize through the four diagnostics in Section~\ref{sec:diagnostic_summary}.

\paragraph{Residual survival, anchor-relative geometry, and task relevance.}
As shown in Table~\ref{tab:cmu_motion}, both Walk and Run exhibit residual survival ($\kappa_r:0.124/0.348$), high excess transverse energy ($\Delta\rho^\perp:60.6/81.7\%$), and improved held-out MSE, respectively.
Without adaptive control of equivariance relaxation, our model achieves performance comparable to EGNO+ACE~\cite{main_ACE_manolache2025learning}, an adaptive equivariance-relaxation approach.
CMU exhibits a surviving residual that is strongly transverse to local anchor variation and improves predictive performance.

\begin{figure}[t]
  \centering
  \includegraphics[width=1\linewidth]{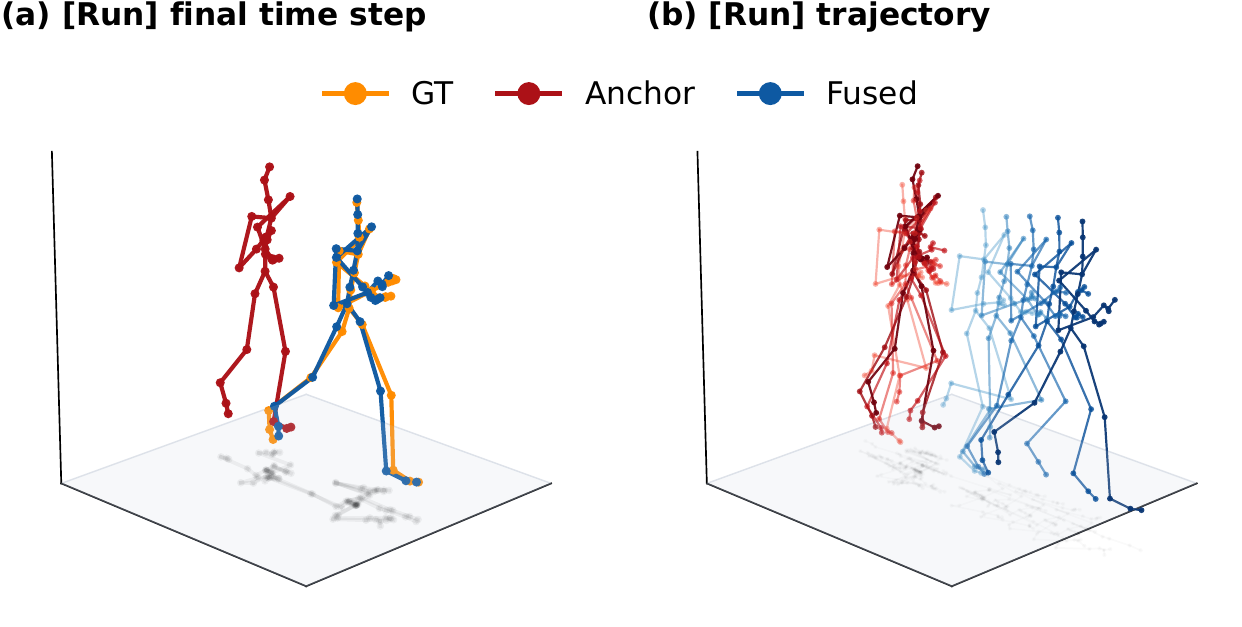}
  \caption{
    CMU Run example.
    (a) The anchor predicts narrower leg separation than the ground truth.
    (b) The anchor shows little forward progression, whereas the fused prediction
    recovers the running trajectory.
    }
  \label{fig:main_CMU_direct}
\end{figure}
\begin{table}[t]
\centering
\caption{
CMU motion results.
Baseline results follow
TFN~\cite{expegnoTFNthomas2018tensor},
SE(3)-Transformer~\cite{expegnose3fuchs2020se},
EGNN~\cite{relatedworkEGNNsatorras2021n},
EGNO~\cite{main_EGNO_xu2024equivariant}, and
EGNO+ACE~\cite{main_ACE_manolache2025learning};
EGNO is the equivariant anchor.
MSE is reported in units of $10^{-2}$.
Local-SVD uses the globally centered trajectory representation
with $K=32$, $\delta=.95$, and $H=32$.}
\label{tab:cmu_motion}
\begin{tabular}{lcc}
\toprule
Method / Metric & Sub \#35 (Walk) & Sub \#9 (Run) \\
\midrule
TFN
    & 32.0 $\pm$ 1.8 & 56.6 $\pm$ 1.7 \\
SE(3)-Tr.
    & 31.5 $\pm$ 2.1 & 61.2 $\pm$ 2.3 \\
EGNN
    & 28.7 $\pm$ 1.6 & 50.9 $\pm$ 0.9 \\
\midrule
EGNO
    & 8.1 $\pm$ 1.6 & 33.9 $\pm$ 1.7 \\
EGNO + ACE
    & 7.4 $\pm$ 0.2 & 23.8 $\pm$ 1.5 \\
\midrule
EGNO + ours
    & \textbf{6.97 $\pm$ 0.90}
    & \textbf{22.83 $\pm$ 2.04} \\
\midrule
$\kappa_r$
    & 0.124 & 0.348 \\
$\Delta\rho^\perp$ (\%) $\uparrow$
    & \textbf{60.6} & \textbf{81.7} \\
\bottomrule
\end{tabular}
\end{table}
\begin{figure}[t]
  \centering
  \includegraphics[width=1\linewidth]{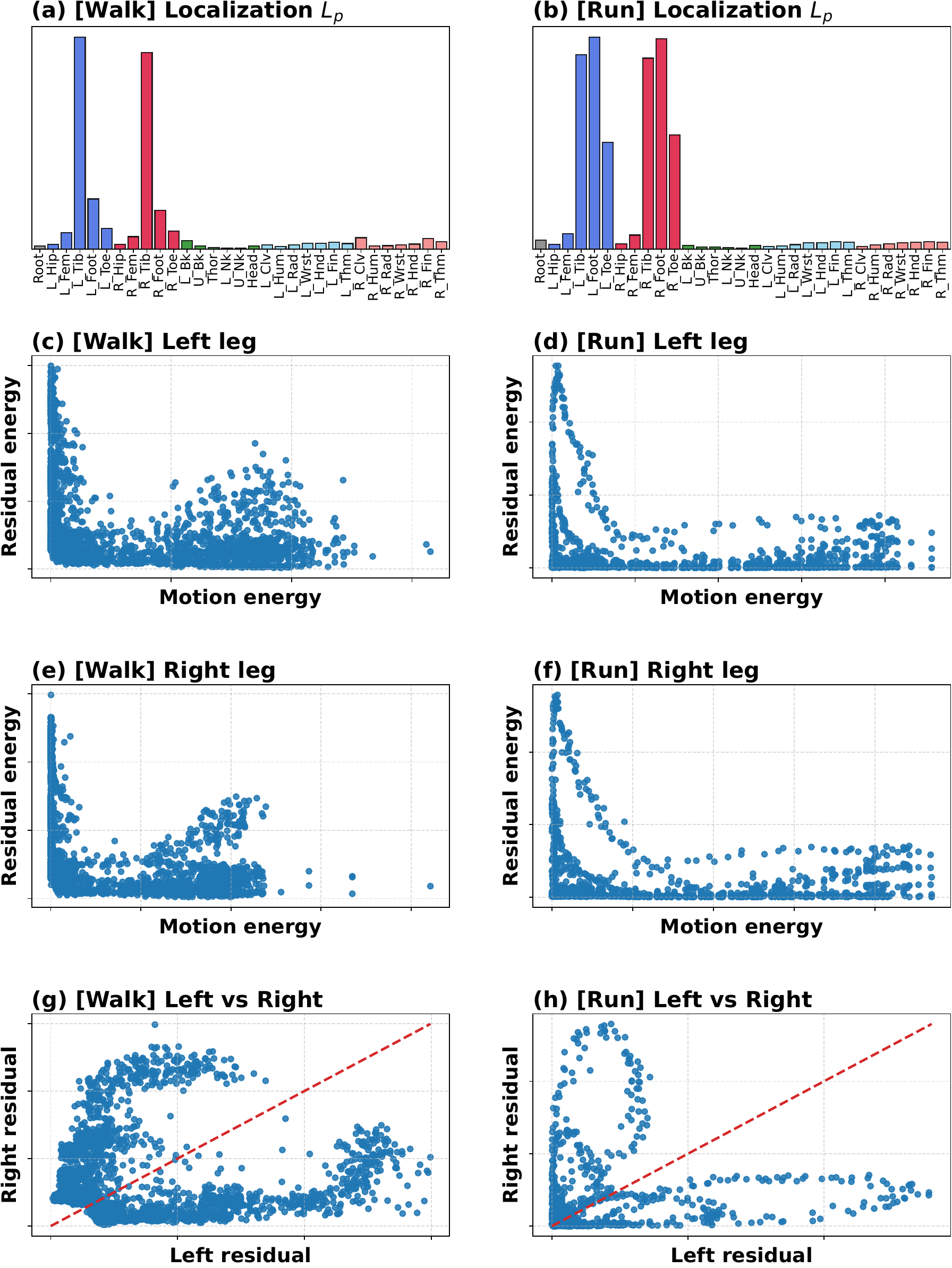}
  \caption{
    CMU physical tracing.
    (a,b) Joint-wise localization $L_p$.
    (c--f) Motion-conditioned residual patterns.
    (g,h) Left--right residual-energy relations.
    }
  \label{fig:main_CMU_physical}
\end{figure}
\paragraph{Direct physical tracing.}
Figure~\ref{fig:main_CMU_direct} shows that the anchor predicts a largely in-place running motion, resembling motion without effective ground propulsion, whereas the fused prediction recovers forward progression.
Thus, a prominent component of the surviving residual compensates for forward translational motion that is underrepresented by the equivariant anchor.
To examine the joint-specific structure accompanying this translational compensation, we apply the direct element-wise diagnostic $L_p$ in Section~\ref{sec:diagnostic_physical_task}, as joint-wise correspondence is preserved at the readout.
The numerator of $L_p$ subtracts the mean residual across joints, isolating joint-specific deviations from any shared translational component.
Figure~\ref{fig:main_CMU_physical}(a,b) shows that the resulting $L_p$ localization is concentrated on the bilateral tibia, foot, and toe joints.
This lower-leg emphasis appears in both motions, but is sharper in Run than in Walk.

However, $L_p$ aggregates over prediction time steps, obscuring the temporal relationship between the localized residual and lower-leg motion.
We therefore quantify lower-leg motion and joint-specific residual variation at each prediction time step using motion and residual energies, respectively.
We denote by $v^{\mathrm{GT}}_{i,t,p}\in\mathbb{R}^3$ the ground-truth velocity of joint $p$, and by $r_{i,t,p}\in\mathbb{R}^3$ its residual vector, for sample $i$ at prediction time step $t$.
Let $S_L$ and $S_R$ denote the tibia, foot, and toe joints of the left and right lower legs, respectively.
For each side $b\in\{L,R\}$, we define
\[
E^{\mathrm{mot}}_{i,t,b}
=
\sum_{p\in S_b}\|v^{\mathrm{GT}}_{i,t,p}\|_2^2,
\quad
E^{\mathrm{res}}_{i,t,b}
=
\sum_{p\in S_b}\|r_{i,t,p}-\bar r_{i,t}\|_2^2,
\]
where $P$ is the total number of body joints and $\bar r_{i,t}=P^{-1}\sum_{p=1}^{P}r_{i,t,p}$ is the mean residual across joints.

In Run, residual energy is largest when lower-leg motion is small in Figure~\ref{fig:main_CMU_physical}(d,f), and the left- and right-leg residual energies exhibit alternating dominance in Figure~\ref{fig:main_CMU_physical}(h).
These patterns suggest stronger residual contributions around ground-contact phases, consistent with alternating ground support during running.
Because ground contact depends on the fixed ground as an external reference beyond the body configuration, we interpret this pattern as consistent with symmetry-breaking structure relative to the prescribed $E(3)$ prior.
This interpretation is qualitatively consistent with prior contact-aware human-motion forecasting \cite{main_CMU_mao2022contact}.
In Walk, this motion--residual pattern is weaker in Figure~\ref{fig:main_CMU_physical}(c,e,g).
Taken together, the recovery of forward progression, residual localization to lower-leg joints near the ground, and motion-dependent patterns consistent with ground-contact phases suggest that task-relevant, ground-associated symmetry-breaking structure can survive through the restricted residual path.

\subsection{ScanObjectNN}
\label{sec:ScanObjectNN}
We evaluate ScanObjectNN \cite{main_scanobjectNN_uy2019revisiting}, a real-world point cloud object classification benchmark, using the challenging PB\_T50\_RS setting and VN-DGCNN \cite{relatedworkDGCNNwang2019dynamic,main_VN_deng2021vector} as the anchor under the same setting as \cite{relatedworkmelnyk2024tetrasphere}.
We compare two matched rotation conditions: $Z/Z$ applies random $z$-axis rotations during both training and testing, preserving the gravity-aligned vertical axis, whereas $\mathrm{SO}(3)/\mathrm{SO}(3)$ applies random 3D rotations during both, largely removing this gravity-related orientation structure.
This comparison tests whether the surviving residual structure depends on the availability of gravity-aligned orientation structure.
ScanObjectNN exhibits asymmetry tunneling, which we characterize through the four diagnostics in Section~\ref{sec:diagnostic_summary}.

\paragraph{Residual survival, anchor-relative geometry, and task relevance.}
As shown in Table~\ref{tab:scanobjectnn}, both $Z/Z$ and $\mathrm{SO}(3)/\mathrm{SO}(3)$ exhibit residual survival ($\kappa_r:0.464/0.177$), high excess transverse energy ($\Delta\rho^\perp:84.0/82.6\%$), and improved held-out accuracy, respectively.
ScanObjectNN exhibits a surviving residual that is strongly transverse to local anchor variation and improves classification accuracy under both rotation conditions.

\begin{figure}[t]
  \centering
  \includegraphics[width=1\linewidth]{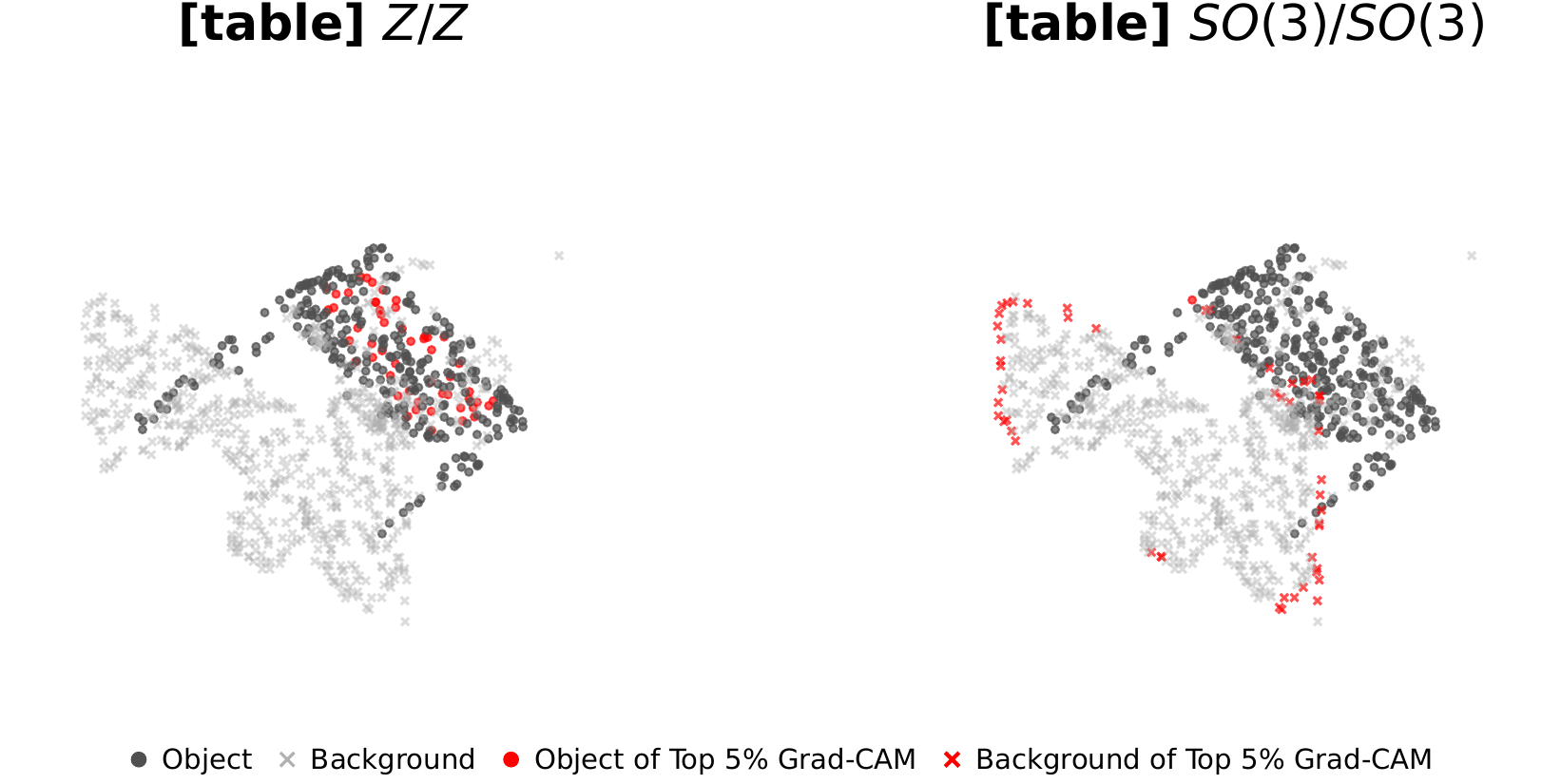}
   \caption{
Residual Grad-CAM visualization on ScanObjectNN under
$Z/Z$ (left) and $\mathrm{SO}(3)/\mathrm{SO}(3)$ (right).
The label \texttt{[table]} denotes the object category.
Red highlights the top $5\%$ attributed points;
circles and crosses denote object and background points.
}
  \label{fig:main_scanobj}
\end{figure}

\paragraph{Attribution-based physical tracing.}
Because global pooling in point-cloud classification removes explicit point-level correspondence at the readout, we apply the Grad-CAM Lift diagnostic in Section~\ref{sec:diagnostic_physical_task} to quantify the localization of anchor and residual attributions to object and background points using the dataset's object/background labels.
As shown in Table~\ref{tab:scanobjectnn}, the anchor remains background-biased relative to the chance-level Lift of $1$ under both rotation conditions.
In contrast, the residual reverses its localization: Object/Background Lift changes from $1.10/0.95$ under $Z/Z$ to $0.94/1.14$ under $\mathrm{SO}(3)/\mathrm{SO}(3)$.
Figure~\ref{fig:main_scanobj} qualitatively illustrates this shift from object-side localization under $Z/Z$ to background-side localization under $\mathrm{SO}(3)/\mathrm{SO}(3)$.

These patterns suggest stronger residual localization to object points under $Z/Z$, aligning residual attribution with the classification target.
Because $Z/Z$ preserves a consistent gravity-aligned orientation reference that can aid object classification, whereas $\mathrm{SO}(3)/\mathrm{SO}(3)$ disrupts this reference, we interpret this rotation-dependent localization as consistent with symmetry-breaking structure relative to the prescribed $\mathrm{SO}(3)$ prior.
Together with the larger accuracy gain under $Z/Z$, these results suggest that task-relevant, gravity-related symmetry-breaking structure can survive through the restricted residual path.
\begin{table}[t]
\centering
\caption{
ScanObjectNN results.
Accuracy baselines follow
\citet{expscanlili2021closer},
\citet{expscanyuyu2023rethinking},
VN-DGCNN~\cite{relatedworkDGCNNwang2019dynamic,main_VN_deng2021vector},
and TetraSphere~\cite{relatedworkmelnyk2024tetrasphere};
VN-DGCNN is the equivariant anchor.
Local-SVD uses $K=128$, $\delta=.95$, and $H=32$.
Lift uses the top-$5\%$ Grad-CAM support and is reported as
Object/Background (Obj/Back), with $1$ denoting chance.
}
\label{tab:scanobjectnn}
\begin{tabular}{lcc}
\toprule
Method / Metric & Z/Z & SO(3)/SO(3) \\
\midrule
\multicolumn{3}{l}{\textit{Accuracy (\%)}} \\
\midrule
Li et al.
    & 74.6 & 74.9 \\
Yu et al.
    & 77.2 & 77.4 \\
VN-DGCNN
    & 77.9 & 78.5 \\
TetraSphere ($K=4$)
    & 79.2 & 79.0 \\
VN-DGCNN + ours
    & \textbf{80.7} & \textbf{79.5} \\
\midrule
\multicolumn{3}{l}{\textit{Residual diagnostics}} \\
\midrule
$\kappa_r$
    & 0.464 & 0.177 \\
$\Delta\rho^\perp$ (\%) $\uparrow$
    & \textbf{84.0} & \textbf{82.6} \\
Anchor Lift (Obj/Back)
    & 0.99 / 1.17 & 0.91 / 1.30 \\
Residual Lift (Obj/Back)
    & 1.10 / 0.95 & 0.94 / 1.14 \\
\bottomrule
\end{tabular}
\end{table}

\subsection{PDBbind}
\label{sec:PDBbind}
\begin{table}[t]
\centering
\setlength{\tabcolsep}{4.0pt}
\caption{
PDBbind results.
GVP denotes the reproduced GVP-GNN baseline
\cite{main_gvp_atom3d_jing2021equivariant}.
Mixed and Had. denote the mixed and Hadamard operators used in the
restricted residual path, respectively.
Local-SVD uses $K=32$, $\delta=.95$, and $H=32$.
Lift uses the top-$5\%$ Grad-CAM support and is reported as
Ligand/Protein (L/P), with $1$ denoting chance.
}
\label{tab:pdbbind}
\begin{tabular}{lccc}
\toprule
Method / Metric & GVP+Mixed & GVP+Had. & GVP \\
\midrule
\multicolumn{4}{l}{\textit{Prediction performance}} \\
\midrule
RMSE $\downarrow$
    & \textbf{1.449} & 1.518 & 1.565 \\
$R$ $\uparrow$
    & \textbf{0.531} & 0.521 & 0.428 \\
\midrule
\multicolumn{4}{l}{\textit{Residual diagnostics}} \\
\midrule
$\kappa_r$
    & 0.223 & 0.298 & -- \\
$\Delta\rho^\perp$ (\%) $\uparrow$
    & \textbf{44.1} & 12.9 & -- \\
Anchor Lift (L/P)
    & 5.00 / 0.77 & 6.35 / 0.69 & -- \\
Residual Lift (L/P)
    & 0.63 / 1.02 & 1.10 / 0.99 & -- \\
\bottomrule
\end{tabular}
\end{table}
We evaluate the ATOM3D experimentally measured ligand-binding affinity (LBA) benchmark derived from the PDBbind refined set \cite{expatom3ddatatownshend2020atom3d,exppdboriwang2004pdbbind,exppdbupdliu2015pdb}, using GVP-GNN \cite{main_gvp_jing2020learning} as the anchor under the original ATOM3D LBA setting \cite{main_gvp_atom3d_jing2021equivariant}, which includes chirality-related local geometry.
Molecular chirality refers to the inability of a molecule to be superimposed on its reflection.
A chiral molecule and its reflected counterpart, known as enantiomers, can interact differently with stereoselective protein binding pockets, resulting in different ligand-binding affinities \cite{main_cannotlearnchirality_dumitrescu2024,wang2025enantioselective}.
Although GVP provides reflection-equivariant functions \cite{main_gvp_jing2020learning}, its scalar predictions remain reflection-invariant and thus cannot distinguish globally reflected molecular configurations \cite{main_cannotlearnchirality_dumitrescu2024}.
To capture molecular chirality beyond the GVP anchor, we introduce mixed-product features into the restricted residual path, following prior work on cross- and mixed-product operations \cite{expatom3dmixedshi2026unifying,dunn2026flowmol3}.
Specifically, the scalar triple product $\chi=(v_1\times v_2)\cdot u$, formed from learned combinations $v_1,v_2$ of GVP vector channels and an edge direction $u$, is invariant under proper rotations but changes sign under reflection.
To examine the role of mixed-product geometry, we compare two restricted residual-path variants that differ only in the feature construction: Mixed uses $\chi_{\mathrm{M}}=(v_1\times v_2)\cdot u$, whereas Hadamard uses $\chi_{\mathrm{H}}=(v_1\odot v_2)\cdot u$, where $\odot$ denotes element-wise multiplication.
Unlike the preceding benchmarks, we explicitly design the residual path to capture reflection-sensitive local geometry associated with molecular chirality and evaluate asymmetry tunneling using the four diagnostics in Section~\ref{sec:diagnostic_summary}.

\paragraph{Residual survival, anchor-relative geometry, and task relevance.}
As shown in Table~\ref{tab:pdbbind}, both Mixed and Hadamard exhibit residual survival ($\kappa_r:0.223/0.298$) and improved held-out RMSE and Pearson correlation, while Mixed exhibits greater excess transverse energy ($\Delta\rho^\perp:44.1/12.9\%$).
PDBbind therefore exhibits surviving residuals under both variants, with Mixed showing stronger anchor-relative transverse structure and greater predictive improvement.

\paragraph{Attribution-based physical tracing.}
Because graph-level pooling in affinity prediction removes explicit atom-level correspondence at the readout, we apply the Grad-CAM Lift diagnostic in Section~\ref{sec:diagnostic_physical_task} to quantify the localization of anchor and residual attributions to ligand and protein atoms.
As shown in Table~\ref{tab:pdbbind}, the anchor is strongly ligand-biased, with Ligand/Protein Lift of $5.00/0.77$ for Mixed and $6.35/0.69$ for Hadamard.
In contrast, the Mixed residual shifts attribution toward protein atoms (Ligand/Protein Lift: $0.63/1.02$), whereas Hadamard retains a slight ligand-side preference ($1.10/0.99$).

The anchor's strong ligand-side attribution suggests that its predictions emphasize ligand-related features.
However, stereoselective binding depends not only on ligand geometry but also on its 3D compatibility with the protein binding pocket \cite{wang2025enantioselective}, highlighting the potential relevance of the Mixed residual's protein-side attribution to reflection-sensitive ligand–protein interactions.
Because mixed-product features can encode reflection-sensitive local geometric distinctions in the ligand–protein binding environment that the GVP anchor cannot resolve, we interpret this operator-dependent localization as consistent with symmetry-breaking structure relative to the prescribed $E(3)$ prior.
Together with the stronger anchor-relative transverse structure ($\Delta\rho^\perp=44.1\%$ for Mixed versus $12.9\%$ for Hadamard) and greater affinity-prediction improvement under Mixed, these results suggest that task-relevant, reflection-sensitive symmetry-breaking structure can survive through the restricted residual path.

\section{Discussion and Limitations}

\paragraph{Cross-task patterns.}
Across the five benchmarks, the four diagnostics in Section~\ref{sec:diagnostic_summary} reveal distinct patterns of residual collapse and survival under strictly equivariant anchors.
Charged N-body, a synthetic symmetry-matched control, exhibits residual collapse without predictive improvement.
ETH/UCY shows weak survival or collapse across subsets, with inconsistent performance changes.
CMU motion exhibits residual survival with high excess transverse energy, predictive gains comparable to those of the adaptive equivariance-relaxation method EGNO+ACE, and ground-associated residual patterns.
In ScanObjectNN, residuals survive with high excess transverse energy under both rotation settings, while gravity-related orientation structure under $Z/Z$ accompanies stronger alignment of residual attribution with the classification target and larger predictive gains.
In PDBbind, residuals survive under both operators, but Mixed, explicitly designed to capture reflection-sensitive geometry associated with molecular chirality, exhibits a shift toward protein-side attribution, higher excess transverse energy, and greater predictive improvement than the Hadamard variant.
Taken together, these observations characterize asymmetry tunneling as a task- and condition-dependent empirical phenomenon.

\paragraph{Scope and limitations of residual diagnosis.}
A natural concern is that predictive gains may simply reflect the residual path's additional capacity rather than symmetry-breaking structure.
Our objective, however, is to characterize asymmetry tunneling as an empirical phenomenon: we design the residual path with ample expressive capacity to represent potential symmetry-breaking structure and examine its transmitted contribution through the four diagnostics in Section~\ref{sec:diagnostic_summary}.
These diagnostics do not directly establish a particular symmetry-breaking mechanism: Local-SVD does not recover an exact manifold-normal component, and physical tracing does not establish causality.
A further limitation arises when the prescribed symmetry group $G$ is misaligned with the task. In this case, residual survival may reflect compensation for an inadequate symmetry prior rather than localized symmetry-breaking structure, while residual collapse does not establish the absence of such structure.

\section{Conclusion}

Across five benchmarks, we observe both residual collapse and asymmetry tunneling under a restricted interface with strictly equivariant anchors, and characterize the surviving residuals through complementary empirical diagnostics.
Our results suggest that a restricted residual path can transmit task-relevant contributions associated with potential symmetry-breaking structure without adaptively controlling equivariance relaxation toward an appropriate level for the data.

\paragraph{Broader perspectives.}
The physical tracing results reveal task-specific localization patterns, while PDBbind illustrates how residual features can be explicitly designed to probe reflection-sensitive molecular geometry.
Together, these observations motivate residual paths that combine ample expressive capacity with physical traceability for investigating potential symmetry-breaking structure while retaining strictly equivariant anchors.

\newpage

\bibliography{icml_2027}
\bibliographystyle{icml2026}

\newpage
\appendix
\onecolumn

\section{Task-Specific Residual Token Correspondence}
\label{app:token_structure}
The residual decoder operates on layer-aligned token representations, but the tokenization depends on the task. For explicitly indexed physical elements, token identity is preserved directly. For unordered point sets and molecular graphs, shared learned queries first aggregate element features into latent tokens. Within the residual decoder, the detached final-layer token representations then serve as cross-attention queries over the layer–token history. The decoded residual contributions are mapped back to physical elements before fusion, preserving the correspondence needed for physical tracing.

In ETH/UCY, predictions are agent-wise, although the ordering of pedestrians has no intrinsic meaning.
The residual representation therefore maintains agent-level correspondence without assigning semantic significance to arbitrary agent ordering.

In CMU motion forecasting, joints and prediction time steps have fixed physical and temporal identities.
The detached anchor states are organized into joint--time tokens, allowing the residual decoder to aggregate layer history while preserving the assignment of each contribution to its corresponding joint and time step.

In ScanObjectNN, unordered points are aggregated into latent tokens using shared learned queries across anchor layers.
The decoded residual contributions are mapped back to individual points through attention-based assignments before global pooling.
This supports point-level localization without treating the input point order as semantically meaningful.

In PDBbind, molecular nodes are exchangeable in their indexing, while graph connectivity and local geometric relationships remain essential.
Node representations, including edge-derived geometric information, are aggregated into latent tokens.
The decoded residual contributions are then mapped back to atoms before graph-level pooling, allowing physical tracing at the node level.

These task-specific constructions preserve the correspondence needed for residual tracing without imposing a universal tokenization rule across different physical structures.

\section{Code Availability}
\label{app:code_availability}
The implementation and experimental code will be made
publicly available at \url{https://github.com/Jeong-Eunho}.


\end{document}